\documentclass[a4paper, 10pt, conference]{ieeeconf}

\newif\ifarwfinalcopy

\arwfinalcopytrue  

\IEEEoverridecommandlockouts                       

\usepackage{graphics} 
\usepackage{epsfig} 
\usepackage{mathptmx} 
\usepackage{times} 
\usepackage{amsmath} 
\usepackage{amssymb}  
\usepackage{bm} 
\usepackage[mathcal]{euscript}  
\usepackage{subfig} 

\usepackage{lineno}
\usepackage{xcolor}
\definecolor{darkgreen}{rgb}{0.0, 0.5, 0.0}
\usepackage{tikzpagenodes}
\usepackage{background}
\usepackage{hyperref}  
\usepackage{graphicx}
\usepackage{tikz}

\ifarwfinalcopy
\backgroundsetup{color=white}
\else

\linenumbers

\newcommand{\MyARWConfidentialLogo}{
\begin{tikzpicture}[remember picture,overlay]
\node[align=center,text=blue] at ([yshift=1em]current page text area.north) {\Large \#\#\# ARW 2025 SUBMISSION: CONFIDENTIAL REVIEW COPY \#\#\#};
\end{tikzpicture}
}

\SetBgContents{\MyARWConfidentialLogo}
\SetBgPosition{current page.north west}
\SetBgOpacity{0.5}
\SetBgAngle{0.0}
\SetBgScale{1.0}

\fi

\title{\LARGE \bf
Multi-Waypoint Path Planning and Motion Control for Non-holonomic Mobile Robots in Agricultural Applications
}

\author{Mahmoud Ghorab and Matthias Lorenzen
\thanks{This work was supported by the BMBF, Deutsche Agentur für Transfer und Innovation within the program DATIpilot.}
\thanks{Mahmoud Ghorab and Matthias Lorenzen are with Institute for Applied Artificial Intelligence and Robotics (IKR), Kempten University of Applied Sciences, Bahnhofstraße 61, 87435 Kempten (Allgäu), Germany {\tt\small \{mahmoud.ghorab, matthias.lorenzen\}@hs-kempten.de}}%
}

\hypersetup{colorlinks=true, linkcolor=blue, citecolor=red, urlcolor=blue}  

\begin{document}

\maketitle

\begin{abstract}
    There is a growing demand for autonomous mobile robots capable of navigating unstructured agricultural environments.
    Tasks such as weed control in meadows require efficient path planning through an unordered set of coordinates while minimizing travel distance and adhering to curvature constraints to prevent soil damage and protect vegetation.
    This paper presents an integrated navigation framework combining a global path planner based on the Dubins Traveling Salesman Problem (DTSP) with a Nonlinear Model Predictive Control (NMPC) strategy for local path planning and control.
    The DTSP generates a minimum-length, curvature-constrained path that efficiently visits all targets, while the NMPC leverages this path to compute control signals to accurately reach each waypoint.
    The system's performance was validated through comparative simulation analysis on real-world field datasets, demonstrating that the coupled DTSP-based planner produced smoother and shorter paths, with a reduction of about 16\% in the provided scenario, compared to decoupled methods.
    Based thereon, the NMPC controller effectively steered the robot to the desired waypoints, while locally optimizing the trajectory and ensuring adherence to constraints.
    These findings demonstrate the potential of the proposed framework for efficient autonomous navigation in agricultural environments.
\end{abstract}


\section{Introduction}
Autonomous navigation in unstructured agricultural environments, such as meadows, poses significant challenges due to unpredictable terrain, the non-holonomic system dynamics of many mobile robots, and the possible presence of both static and dynamic obstacles~\cite{mammarellaCooperationUnmannedSystems2022a}.
An ecological weed control system is a prime application where efficient navigation is crucial, enabling the reduction of herbicide use and minimizing human intervention.
In the considered application, the process begins by selecting a geo-fence that defines the field's safety boundaries, ensuring the robot operates within a designated area.
Next, the target weeds are autonomously detected and mapped during a scanning phase.
Once the scanning and mapping are complete, the robot is tasked with navigating to the identified weeds and eliminating them using a mechanical weed removal tool, avoiding the use of chemical herbicides.
This last phase is the primary focus of this work, where the proposed DTSP-based global path planner, as well as the NMPC local path planner and waypoint-following controller are integrated to efficiently guide the robot to each detected weed, while considering the different robot and environmental constraints.

However, the order in which the targets should be visited is not determined a priori.
Hence, the objective of the global path planner is to generate a feasible path of minimum length that efficiently visits all the targets.
%
The problem of determining the order of waypoints to minimize travel distance is typically formulated as an Euclidean Traveling Salesman Problem (ETSP).
However, solving the ETSP alone does not consider the vehicle's non-holonomic constraints or environmental constraints, such as avoiding damage to soil and healthy grass by preventing arbitrarily sharp turns in the path.
Therefore, the planner has to consider curvature constraints, by generating a minimum length path making use of Dubins curves instead of straight line segments.
This formulation, known as the Dubins Traveling Salesman Problem (DTSP), extends the classical ETSP to non-holonomic vehicles with a minimum turning radius constraint~\cite{savlaPointtopointTravelingSalesperson2005}.
%

While the DTSP based planner provides a feasible global path to guide the robot towards each waypoint, a local planner and controller is essential for ensuring safe and adaptive navigation in dynamic environments and computing the necessary control input.
The presence of static obstacles and dynamic agents, such as animals, human workers or other robots operating in the field, requires real-time local path replanning.
To this end, Nonlinear Model Predictive Control (NMPC) is employed as both the local path planner and waypoint following controller within the same framework.


\subsection{Related Work}
\label{subsec:related_work}
Various formulations and extensions of the DTSP and NMPC have been presented in the literature, each with its own advantages and trade-offs.
Selecting the right combination of DTSP and NMPC formulations is crucial for achieving efficient and reliable navigation in agricultural environments.
The choice directly impacts the optimality of the generated paths, the overall motion control objectives, and the ability to meet specific task requirements while adhering to overall system’s constraints.

Approaches to solving DTSP primarily differ in how they determine the ordering of waypoints and compute the associated orientations. These differences influence the accuracy of the near optimal solution, and the computational effort.
Similarly, NMPC formulations vary in terms of cost function design, constraints handling, and real-time performance, making the selection process highly application-dependent.

This work emphasizes the importance of choosing the most suitable DTSP and NMPC formulations tailored to agricultural applications, balancing global path feasibility, motion planning adaptability, and considering real-world operational constraints.

\subsubsection{DTSP-based Global Path Planning}
\label{subsubsec:dtsp}

In~\cite{dubinsCurvesMinimalLength1957} Dubins introduced a method for determining the shortest path in a 2D space, given curvature constraints as well as the entry and exit orientations between two points as input.
The resulting path consists of a combination of straight line segments and arcs with radii that adhere to the vehicle's curvature constraints.

The DTSP was first introduced by~\cite{savlaPointtopointTravelingSalesperson2005}. In this extension of the classical TSP, the path connecting any two points must be a Dubins curve and two curves that meet at the same point must share the same orientation.

The core distinctions between methods addressing the DTSP lie in how they determine the ordering of the waypoints and calculate the orientations associated with the points.
Interested readers are referred to the comprehensive survey~\cite{macharetSurveyRoutingProblems2018} for a detailed review of the various routing methods.

Existing literature mostly adopted a decoupled approach for route generation~\cite{savlaPointtopointTravelingSalesperson2005, maRecedingHorizonPlanning2006, rathinamResourceAllocationAlgorithm2007, macharetDataGatheringTour2012}.
Thereby, first, the visiting sequence is determined solving the ETSP.
Then, the vehicle's orientation at each point is defined, for example, using the Alternating Algorithm (AA)~\cite{savlaPointtopointTravelingSalesperson2005}.
Finally, the waypoints are connected with Dubins curves.
However, relying solely on the Euclidean distance metric to define the visit order does not necessarily yield efficient results when using Dubins curves for path generation.
This approach can lead to excessive circular maneuvers, especially in dense waypoint configurations typical of autonomous weed control applications.
Since the optimization of waypoint coordinates and headings is inherently coupled, decoupling them compromises optimality~\cite{vanaOptimalSolutionGeneralized2020}.
As a result, a tour based solely on the ETSP ordering cannot achieve an approximation ratio better than $O(n)$ (i.e., the best solution is within a factor of $n$ of the optimal solution) see~\cite{nyDubinsTravelingSalesman2012}.

In the coupled approach, the sequence is determined by directly using the lengths of the Dubins curves between pairs of points.
However, the main challenge here is to find the right mechanism to determine the entry and exit orientations without even having a predefined sequence of points.
In~\cite{lenyApproximationAlgorithmCurvatureConstrained2010}, the orientations of all points are initially set to zero (or a fixed random value), and all interconnecting curves are calculated and connected to form a complete graph.
An instance of the Asymmetric TSP (ATSP) is then solved to find the shortest path in this graph.
This method was later extended to include a complete heading discretization~\cite{lenyPerformanceOptimizationUnmanned2008}.
The technique involves selecting a finite set of $k$ possible headings at each waypoint.
A graph is created with $n$ clusters, each representing a waypoint and containing $k$ nodes that correspond to different headings.
Subsequently, the Dubins distance between configurations of node pairs from different clusters is computed.
Finally, a tour through all clusters, containing exactly one point per cluster, is then determined.
A logarithmic approximation ratio $O(log(n))$ for this ATSP can be achieved by directly solving the problem using available algorithms implementations, such as those described in~\cite{helsgaunEffectiveImplementationLin2000, friezeWorstcasePerformanceAlgorithms1982, kaplanApproximationAlgorithmsAsymmetric2003}.

In both DTSP formulations and most global planners in general, solutions are computed under tight
time constraints, often resulting in suboptimal paths based on simplified models.
Consequently, there is considerable room for improvement by integrating appropriate motion planning and control systems to further locally optimize the global path.

\subsubsection{NMPC-based Motion Planning}
The fundamental principle of MPC is to use the system's model to forecast its future behavior and optimally adjust control actions by solving a constrained optimization problem over a receding horizon at each sampling time~\cite{rawlingsModelPredictiveControl2017, gruneNonlinearModelPredictive2017}.
By minimizing a cost function that incorporates possible nonlinear multi-input multi-output (MIMO) system dynamics along with state and input constraints, NMPC has proven to be a promising approach for various applications, including stabilization, tracking, and motion planning of mobile robots in unstructured and dynamic environments~\cite{mehrezPredictivePathFollowing2017, britoModelPredictiveContouring2019, solopertoNonlinearMPCScheme2023,lorenzenMPCbasedMotionPlanning2025}.

In automated weed control applications, the primary objective is for the robot to reach and stop at each designated waypoint.
This is ensured by making the state corresponding to the desired pose a stable attractor of the feedback control loop.
A conventional approach to ensure this with NMPC involves enforcing terminal costs and/or terminal region constraints near the desired set-point.
However, when the set-point is located at relatively long distance from the robot, the prediction horizon required becomes prohibitively long for practical applications.
An alternative strategy is to reformulate the problem as one of path following by generating a global path that connects all waypoints and then following this path piece-wise~\cite{faulwasserModelPredictivePathfollowing2009, yuNonlinearModelPredictive2015, mehrezPredictivePathFollowing2017}.

In the considered application, as in many other applications, the goal is to reach the target while satisfying constraints rather than strictly following a specific path. As noted in Section \ref{subsubsec:dtsp}, global planners often yield suboptimal paths when computed in finite time, particularly under kinematic and dynamic constraints. Therefore, exactly following these paths can complicate motion control and make it impossible when real-time obstacle avoidance is required. Instead, a flexible approach that allows the motion planner to dynamically optimize the global path and find shortcuts is preferred.

A novel NMPC formulation, proposed in~\cite{lorenzenMPCbasedMotionPlanning2025}, guarantees convergence to a desired target while ensuring closed-loop stability, adherence to system constraints, and collision avoidance with obstacles.
The method optimally selects an artificially generated reference set-point, dynamically adjusted along the global reference path, which guides the robot without requiring strict path following.
This artificial reference is used to define feasible stabilizing terminal constraints.

This work adapts and integrates the coupled DTSP formulation from~\cite{lenyPerformanceOptimizationUnmanned2008} with the NMPC-based motion planner from~\cite{lorenzenMPCbasedMotionPlanning2025} to enable an automated, robot-based weed control application. The resulting integrated framework addresses a critical gap in applied research by combining a multi-waypoint, curvature-constrained DTSP-based global planner with an advanced NMPC-based local motion planner and controller tailored for agricultural robots.

The remainder of the paper is organized as follows. Section \ref{sec:methods} details the proposed system, explaining the integration of the DTSP-based global path planner with the NMPC-based local path planner and waypoint follower. Section \ref{sec:results} describes the simulation setup and presents a comparative analysis of the results. Finally, Section \ref{sec:Summary_Outlook} concludes the paper and outlines directions for future work.

\section{Proposed Navigation and Control System}
\label{sec:methods}
\subsection{System Overview}
The proposed system integrates a two-layer architecture for autonomous navigation.
The global path planner, based on the coupled DTSP formulation, processes unordered multi-waypoint coordinates to compute an optimal sequence of curvature-constrained Dubins paths connecting these waypoints.
These paths minimize travel distance while adhering to curvature constraints tailored specifically for agricultural applications, where sharp turns can damage the soil and grass.
The NMPC-based local path planning and waypoint following algorithm utilizes the resulting global Dubins path to ensure precise convergence to each waypoint while respecting different system constraints.

\subsection{DTSP Algorithm}
Given $W$ waypoints in a 2D space, the DTSP aims to determine the shortest path that connects all points while adhering to curvature constraints.
Consequently, the path between any two points should be a Dubins curve, and the curves meeting at the same point must share the same orientation.

The following steps present the DTSP routing problem based on~\cite{lenyPerformanceOptimizationUnmanned2008}:
\begin{enumerate}
    \item For each of the $W$ target points, select $K$ candidate headings (e.g., $k\frac{2\pi}{K}$ for $k\in \{0,1,...,K-1\}$).
    \item Represent each target as a cluster of $K$ nodes, where each node corresponds to a configuration $q_i = (p_i,\theta_i)$ with position $p$ and a candidate heading $\theta$. The total number of nodes is $nK$.
    \item For each pair of nodes $q_i$ and $q_j$ that belong to different clusters (i.e., different targets), compute the Dubins curve with minimum distance $\mathcal{D}_\rho(q_i,q_j)$.
          This curve is parameterized by the minimum turning radius $\rho$, defines the cost for traveling from a specific configuration at target $i$ to a different one at target $j$.
    \item Arrange the computed Dubins distances into a cost matrix $M$ of size $N \times N$, where $N = nK$.
\end{enumerate}



From the matrix $M$, one can construct an ordered sequence $\mathbf{Q}_{\Sigma} \;=\;(q_{\Sigma (0)}, q_{\Sigma (1)} , \dots , q_{\Sigma (N-1)} )$ which represent some permutation $\Sigma$ of configurations $q_{\Sigma (i)} = (p_{\Sigma (i)},\theta_{\Sigma (i)})$ of a complete tour of the mobile robot, after excluding transitions between configurations within the same target.

Based on this representation, the corresponding objective function can be formulated as follows:
\begin{equation}
    \underset{\theta, \, \Sigma}{\text{minimize}} \quad \mathcal{L}_{\rho} (\mathbf{Q}_{\Sigma})
    \label{eq:dtsp_obj}
\end{equation}
Where the cost function is defined as:
\begin{equation}   
     \mathcal{L}_{\rho} (\mathbf{Q}_{\Sigma}) = \mathcal{D}_{\rho} (q_{\Sigma (N-1)}, q_{\Sigma (0)}) + \sum_{i=0}^{N-2} \mathcal{D}_{\rho} (q_{\Sigma (i)}, q_{\Sigma (i+1)})
    \label{eq:dtsp_cost}
\end{equation}

\subsection{NMPC Algorithm}




The robot's motion is governed by a discrete-time, nonlinear dynamic system, described by the following difference equation:
\begin{equation}
    \mathbf{x}(n+1) = f(\mathbf{x}(n), \mathbf{u}(n)),
\end{equation}
where \( f : \mathbb{R}^{n_x} \times \mathbb{R}^{n_u} \to \mathbb{R}^{n_x} \) is a continuous function that models the system dynamics.
Here, \( x(n) \in \mathbb{R}^{n_x} \) represents the system state, while \( u(n) \in \mathbb{R}^{n_u} \) denotes the control input at the sampling time $t_n$, where $n = 0, 1, 2, \dots$.

The global path $\mathcal{P}_d$ generated from the DTSP-based planner can be represented as a sequence of path segments connecting each pair of consecutive waypoint poses as follows:
\begin{equation}
    \mathcal{P}_d = (p_0, p_1, \dots, p_{W-1}),
\end{equation}

where $W$ is the total number of waypoints.
Each path segment $p_w$ is described as a continuous function:
\begin{equation}
    p_w : [0, 1] \mapsto \mathbb{R}^{n_{x}},
\end{equation}
where $p_w(0)$ represents the initial configuration of the path segment, while $p_w(1)$ represents the target configuration.

The following NMPC formulation used in this work was originally proposed in~\cite{lorenzenMPCbasedMotionPlanning2025}.
This approach ensures that both constraint satisfaction and convergence to a desired target can be guaranteed.
Unlike traditional path-following approaches, this method does not require the robot to strictly follow the reference path $p_w$.
Instead, the path only serves as a guidance mechanism to identify a suitable terminal constraint, which guarantees that at each control step, the local solution computed by the NMPC algorithm can be suitably extended to reach the target pose.
This is achieved by introducing an artificial reference, which serves as an intermediate target configuration and is optimized within the NMPC optimization problem.

In the following, the predicted state and control input trajectories over the finite prediction horizon \( N \) are denoted as \( \mathbf{\bar{x}}(\cdot) \in X \)  and \( \mathbf{\bar{u}}(\cdot) \in U \), where $X$ and $U$ represent the set of admissible states and inputs respectively. These trajectories are defined as
\begin{align}
    \mathbf{\bar{x}}(\cdot) & = (\mathbf{\bar{x}}(1), \mathbf{\bar{x}}(2), \dots, \mathbf{\bar{x}}(N)),   \\
    \mathbf{\bar{u}}(\cdot) & = (\mathbf{\bar{u}}(0), \mathbf{\bar{u}}(1), \dots, \mathbf{\bar{u}}(N-1)).
\end{align}

The artificial reference is chosen along the current path segment $p_w$. With the additional optimization variable $\bar{s}\in [0, 1]$ and the path $p_w$, this artificial reference is given by $p_w(\bar{s})$.

The MPC cost function is defined by
\begin{equation}
    J_N(\mathbf{x}_0, \mathbf{\bar{x}}(\cdot), \mathbf{\bar{u}}(\cdot), \bar{s}) = \sum_{k=0}^{N-1} \ell(\mathbf{\bar{x}}(k), \mathbf{\bar{u}}(k)) + V_o(\bar{s}),
\end{equation}
where the stage cost $\ell : \mathbb{R}^{n_{x}+n_{u}} \to \mathbb{R}_{\leq 0}$ and offset cost $V_0 : [0, 1] \to \mathbb{R}_{\geq 0}$ are positive definite functions.
We define the stage cost
\begin{equation}
    \ell(\mathbf{\bar{x}}(k), \mathbf{\bar{u}}(k)) = \left\Vert \mathbf{\bar{x}}(k) - p(\bar{s}) \right\Vert^4_{Q} + \left\Vert \mathbf{\bar{u}}(k) \right\Vert^4_{R},
\end{equation}
where \( Q \) and \( R \) are positive definite weighting matrices that penalize the deviation of the predicted states from the intermediate artificial reference pose and penalize excessive control effort, respectively.

The offset cost \( V_o(\bar{s}) \) ensures that the artificial reference progresses forward toward the final target pose \( p_w(1) \)
as it penalizes the distance along the path between the current artificial reference and the target pose.
Is defined by
\begin{equation}
    V_o(\bar{s}) = q_s (1 - \bar{s})^2,
\end{equation}
where \( q_s \) is a positive weighting scalar that penalizes the deviation between the final reference index $1$ and the current optimal intermediate artificial reference $\bar{s}$.

Finally, the NMPC algorithm at each sampling time $t_n$, $n = 0, 1, 2, \dots$, can be described as follows:
\begin{enumerate}
    \item Measure the state $\mathbf{x}(n) \in X$ of the robot.
    \item Set $\mathbf{x}_0 = \mathbf{x}(n)$, solve the optimal control problem (OCP) defined by:
          \begin{subequations}\label{ocp}
              \begin{align}
                  \underset{\mathbf{\bar{u}}(\cdot), \, \bar{s}}{\text{minimize}} & \quad J_N(\mathbf{x}_0, \mathbf{\bar{x}}(\cdot), \mathbf{\bar{u}}(\cdot), \bar{s}) \label{ocp:a} \\[6pt]
                  \text{s.t.} \quad \mathbf{\bar{x}}(0)                           & = \mathbf{x}_0 \label{ocp:b}                                                                     \\
                  \mathbf{\bar{x}}(k+1)                                           & = f\bigl(\mathbf{\bar{x}}(k), \mathbf{\bar{u}}(k)\bigr), \quad k \in [0,\, N-1] \label{ocp:c}    \\
                  \mathbf{\bar{x}}(k)                                             & \in X, \quad k \in [1,\, N] \label{ocp:box_x}                                                    \\
                  \mathbf{\bar{u}}(k)                                             & \in U, \quad k \in [0,\, N-1] \label{ocp:box_u}                                                  \\
                  \mathbf{\bar{x}}(N)                                             & = p(\bar{s}) \label{ocp:terminal_constraints}
                  \\
                  \bar{s}                                                         & \in [0,1] \label{ocp:art_ref}                                                                    \\
                  \mathcal{B}\bigl(\mathbf{\bar{x}}(k)\bigr)\cap\mathcal{O}_i
                                                                                  & = \varnothing, \quad k \in [1,\,N],\; i \in [1,\,N_{o}]
                  \label{ocp:obstaceles_avoidance}
              \end{align}
          \end{subequations}
    \item Denote the obtained optimal solution $\mathbf{u}^*(\cdot)$, $\mathbf{x}^*(\cdot)$, $s^*$.
    \item Apply the control input $\mathbf{u}(n) = \mathbf{{u}}^*(0)$ to the system.
    \item Repeat until the robot reaches the final waypoint, then start over using the next path segment.
\end{enumerate}

General constraints on states and control inputs for nonlinear systems are incorporated into the OCP in the form of set membership conditions, as defined in \eqref{ocp:box_x} and \eqref{ocp:box_u}, respectively.
Furthermore, static obstacle avoidance can be also considered in the optimization problem by considering constraints \eqref{ocp:obstaceles_avoidance}. Where $ \mathcal{B}$ represents the robot's footprint, and $\mathcal{O}_i$ denotes the $i$-th obstacle in the environment.


\section{Results}
\label{sec:results}
The proposed system is evaluated in a simulated agricultural scenario, where a mobile robot navigates to a set of target weeds.
The results are presented in terms of path planning and waypoint-following performance metrics, including path length, target reaching, smoothness, and curvature constraints adherence.
A comparative analysis is conducted between the proposed DTSP planner with angle discretization and the decoupled approach based on the Alternating Algorithm (AA), see Section \ref{subsec:related_work} and~\cite{savlaPointtopointTravelingSalesperson2005}.
The results demonstrate the effectiveness of the integrated global planner and NMPC methods adapted in this work.
\subsection{Simulation Setup}

The simulation scenario consists of a 2D field with a set of target weeds distributed across the area.
In this phase, the global path planner generates a Dubins path that connects all target weeds in the field, while the NMPC controller optimizes the robot's trajectory to reach each detected weed accurately while adhering to constraints from the robot's kinematics and the environment.

After formulating the DTSP and transforming it into an ATSP, the problem was solved using the LKH optimizer, which is an effective implementation of the Lin-Kernighan traveling salesman heuristic~\cite{helsgaunEffectiveImplementationLin2000}.

The NMPC problem is symbolically formulated in MATLAB using the CasADi framework~\cite{anderssonCasADiSoftwareFramework2019}.
To ensure a smooth and continuously differentiable path function, the global Dubins reference path is first sampled at 5 cm intervals and then converted into a CasADi function, $p(s)$, using CasADi's linear interpolation utilities.
This function is parameterized over the normalized domain $s \in [0,1]$.

In this agricultural application a differential-driven mobile robot model as described in~\cite{siegwartIntroductionAutonomousMobile2004} is utilized:
\begin{equation}
    \dot{\mathbf{x}} =
    \begin{bmatrix}
        \dot{x} \\
        \dot{y} \\
        \dot{\theta}
    \end{bmatrix}
    =
    \begin{bmatrix}
        v \cos(\theta) \\
        v \sin(\theta) \\
        \omega
    \end{bmatrix}
\end{equation}
The robot's control inputs are defined as $\mathbf{u} = [v \; \omega]^T$, where $v$ and $\omega$ represent the linear and angular velocity respectively.
The output states of the robot are given by $\mathbf{x} = [x \; y \; \theta]^T$, which represent the 2D pose of the robot, including its position $(x, y)$ and orientation $\theta$.
This mathematical model is employed for both the simulation and prediction models, without taking into account possible process or measurement noise.



The prediction model is integrated using the fourth-order Runge-Kutta (RK4) method to compute the state evolution over each discretization interval. The continuous-time OCP is discretized via direct multiple shooting, which converts it into a nonlinear programming (NLP) problem that is then solved with the Interior Point Optimizer (IPOPT)~\cite{wachterImplementationInteriorpointFilter2006}.

The NMPC problem is parameterized by a sampling time of \(\Delta t = 0.1\) seconds and a prediction horizon of \(N = 20\). The weight matrices are defined as
\begin{equation*}
    \begin{aligned}
        Q   & = \operatorname{diag}(0.1,\ 0.1,\ 0.01),\, \\
        R   & = \operatorname{diag}(0.1,\ 1.0),\,        \\
        q_s & = 10^{4}.
    \end{aligned}
\end{equation*}
The minimum turning radius constraint, required in this application, is enforced by the inequality constraint
\begin{equation*}
    \bar{v}(k) \geq r_{\text{min}} |\bar{\omega}(k)|
\end{equation*}
which is added to the optimal control problem.
Furthermore, control inputs box constraints
\[
    \mathbf{u}_{min} \leq  \mathbf{\bar{u}}(k)  \leq
    \mathbf{u}_{max}
\]
are taken into account to limit the robot's linear and angular velocity.
Finally, to ensure smooth motion, in such agricultural application it is convenient to also consider constraints on the rate of change of control inputs (i.e., acceleration of the robot)
\[
    \Delta \mathbf{u}_{min} \leq  \mathbf{\bar{u}}(k) - \mathbf{\bar{u}}(k-1)  \leq
    \Delta \mathbf{u}_{max}.
\]
The robot considered in this work has maximum linear velocity of 0.5 m/s and a maximum angular velocity of 1.9 rad/s. The rate of change constraints are defined as a fraction of the maximum control values, allowing adaptation based on operational requirements (e.g., $\mathbf{u}_{max}/n$), where $n \in [1,\,N_{o}]$
\subsection{Simulation Results}

\begin{figure*}[thpb]
    \centering
    \subfloat[Reference path generated using the proposed coupled formulation of the DTSP-based planner with 10 angle discretization levels.
        This method eliminates loops and provides a more efficient path.]{
        \includegraphics[width=0.45\textwidth]{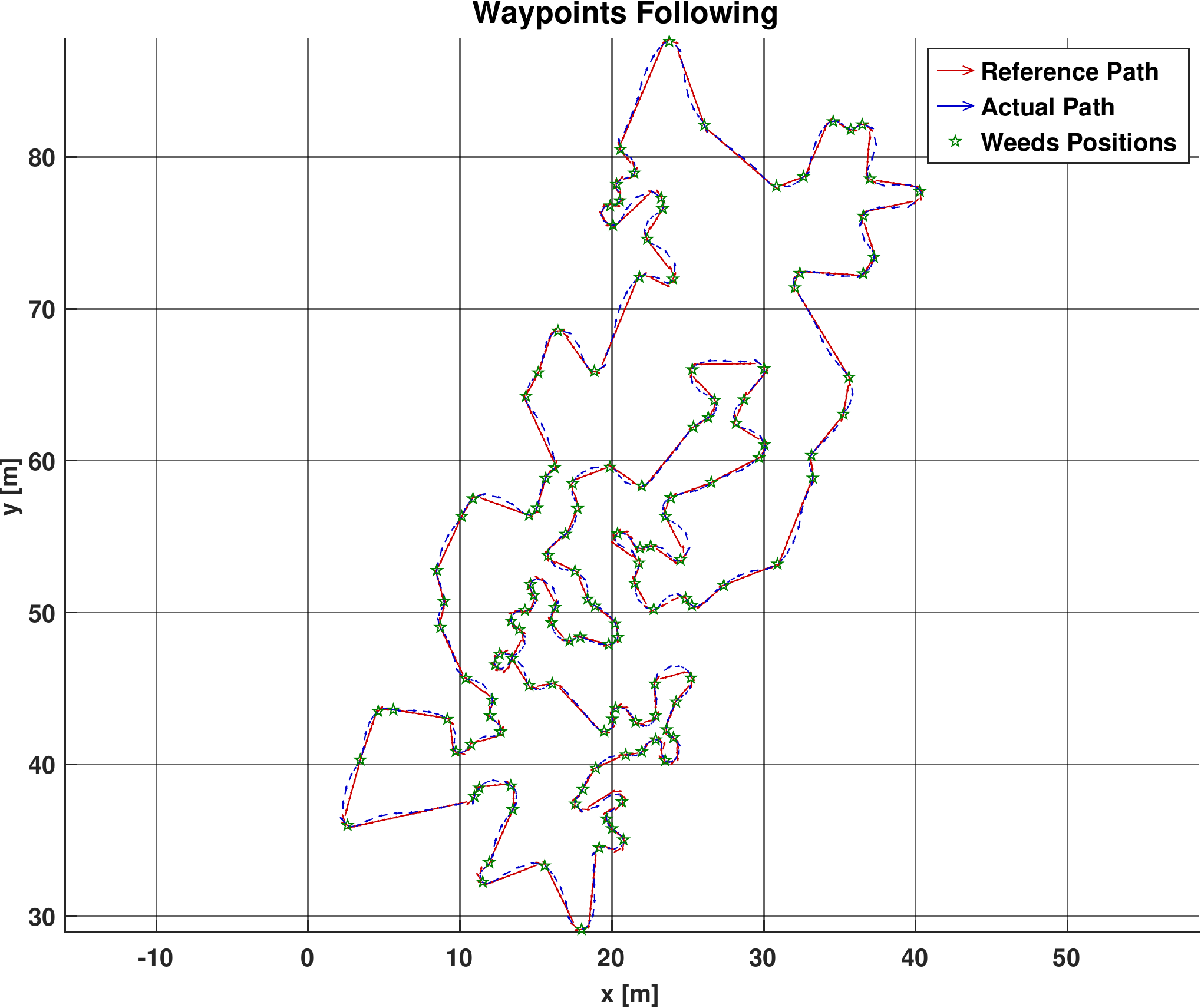}
        \label{fig:case_a}
    }
    \hfill
    \subfloat[Reference path generated using a decoupled formulation of the DTSP-based planner with the Alternating Algorithm for angle selection.
        This approach results in suboptimal paths with redundant circular loops.]{

        \includegraphics[width=0.45\textwidth]{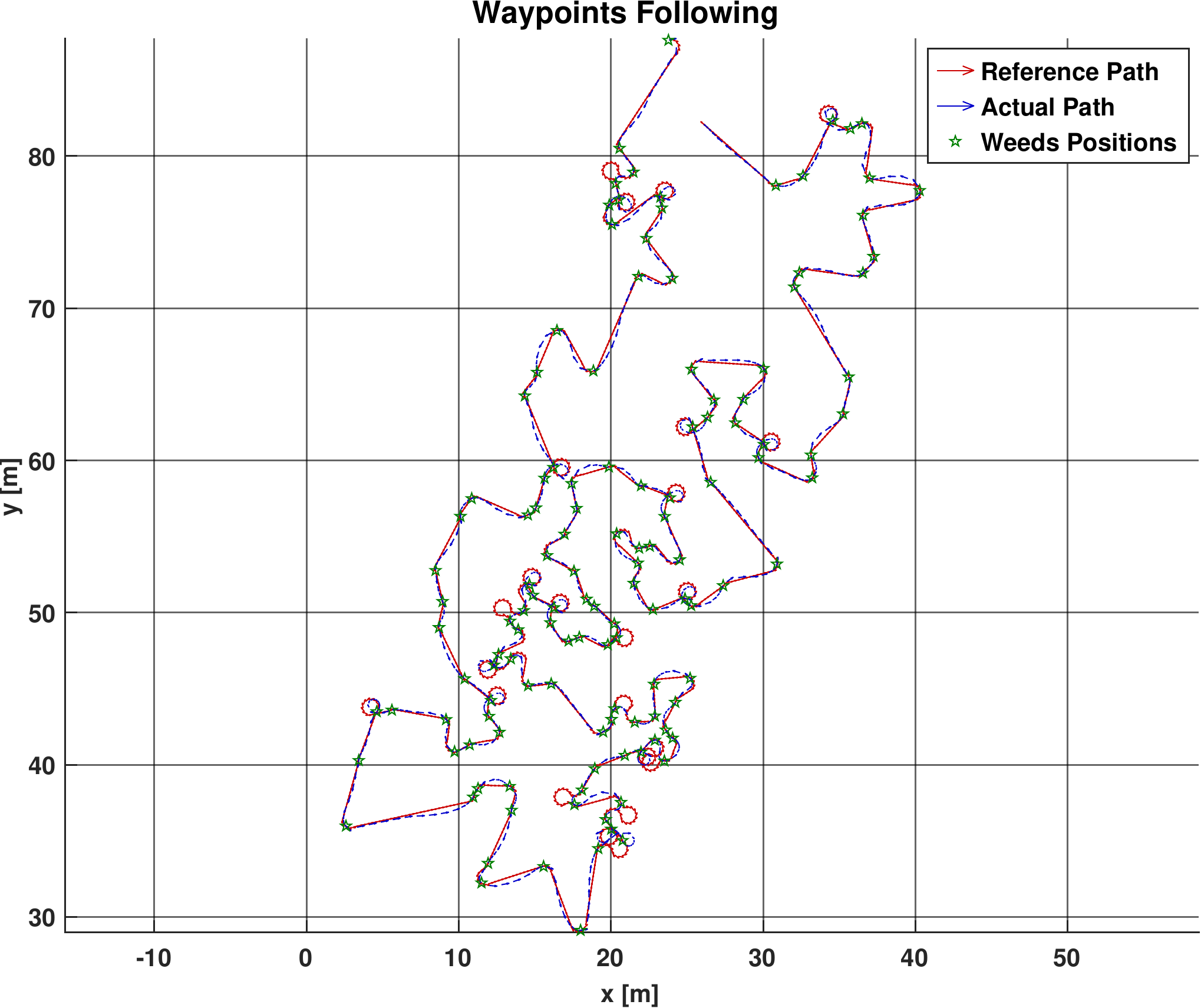}
        \label{fig:case_b}
    }
    \caption{
        Comparison of Dubins tours (red arrows) for approximately 20~m~$\times$~60~m field containing 150 target weeds (green stars), with a vehicle turning radius of 0.5~m.
        The proposed coupled DTSP-based planner \ref{sub@fig:case_a} chooses a different order of the waypoints, thereby allowing for a smoother path, whereas the decoupled DTSP-based planner \ref{sub@fig:case_b} results in a less optimal path with redundant loops.
        In both cases, the NMPC closed-loop state trajectory (blue arrows) successfully reaches all waypoints while locally optimizing motion by smoothing sharp turns and taking efficient shortcuts when beneficial.
    }
    \label{fig:dtsp_compare}
\end{figure*}

The test scenario illustrated in Fig. \ref{fig:dtsp_compare} evaluates the performance of the proposed DTSP global path planner (Fig. \ref{fig:case_a}) against a DTSP planner from the decoupled category (Fig. \ref{fig:case_b}), as discussed in Section \ref{subsec:related_work}.
This planner utilizes the Alternating Algorithm (AA) to determine the waypoints orientations, whereas the DTSP method applied in this work incorporates 10 angle discretization levels for each waypoint.
Both planners were tested on the same dataset, consisting of 150 target weeds distributed across approximately 20$\times$60 square meters field, with a vehicle turning radius constraint of 0.5 meters.

In both cases, the proposed NMPC algorithm was able to optimize the reference paths and accurately reach each waypoint, while still respecting the turning radius constraints required to protect the soil and grass from damage.
A steady-state error of no more than 0.05 meters was achieved at each target pose.

The proposed DTSP planner presented in Fig.~\ref{fig:case_a}, achieved a total path length of 323.49 meters, outperforming the decoupled approach shown in Fig.~\ref{fig:case_b}, which resulted in a path length of 384.58 meters, i.e. nearly 19\% longer.
For reference, the shortest possible path computed by only solving the ETSP without considering curvature constraints was 314.20 meters.

As observed in Fig. \ref{fig:case_b}, the path generated by the DTSP planner using the alternating algorithm is suboptimal, characterized by numerous loops that are necessary to reach the next waypoint given the curvature constraints.
In contrast, the proposed DTSP planner with 10 angle discretization levels, as shown in Fig. \ref{fig:case_a}, leads to a different order of the waypoints, allowing for a significantly smoother path.
This path connects all targets with hardly any redundant loops, which can effectively guide the NMPC towards the targets.
Experiments with angle discretization, starting from three orientations per waypoint and incrementally increasing, showed that higher discretization levels generally reduced path cost but also increased computational time. This trade-off depends on factors such as the density of targets and the turning-radius constraints.
Furthermore, the benefits of using a coupled approach quickly grow with a higher target density and a larger minimum turning radius.

As depicted in Fig. \ref{fig:dtsp_compare}, the robot successfully navigates all target weeds accurately while adhering to curvature constraints.
Thereby the proposed NMPC does not strictly follow the reference path but locally optimizes the trajectory based on the NMPC cost function.
E.g., to protect the soil, tight turns are discouraged, leading to wider turns to smooth out tight turns from the global planner, as long as this does not significantly increase the path length.
On the other hand, it takes shortcuts by making tighter turns when this helps to significantly reduce the travel distance.
This local planning behavior of the NMPC can be tuned by adjusting the prediction horizon length, the cost function weights, and the allowable turning radius.


\section{Summary and Outlook}
\label{sec:Summary_Outlook}
This paper has presented a practical autonomous navigation framework for non-holonomic mobile robots in agricultural applications.
Given target coordinates, the proposed framework integrates a global path planner based on a coupled DTSP formulation with an NMPC-based motion planning and control strategy to generate feasible reference paths and compute optimal control inputs that satisfy both the robotic system constraints and the operational demands of the agricultural environment.

The system's performance was validated through a comparative analysis with a reference path generated by a global planner based on a decoupled DTSP formulation, demonstrating the advantages of the applied DTSP approach and its effectiveness as a reference input for the local motion planner and controller.
By optimally selecting a feasible artificial reference and corresponding terminal constraint along the planned path, the NMPC methodology smooths out sharp turns, identifies efficient shortcuts, and ensures precise waypoint navigation while maintaining overall system stability under various constraints.

Future research will focus on enhancing local motion planning by considering complex obstacle scenarios, including moving humans, animals, other robots and machinery into the NMPC's OCP formulation for safe, real-time adaptation to moving agents.
Experimental field validation is planned under varying terrain conditions to address challenges arising from process and measurements noise, bridging the gap between simulation and practical agricultural robotics.
{\small
      \bibliographystyle{IEEEtranS}
      \bibliography{ML_MG}

\begin{thebibliography}{10}
\providecommand{\url}[1]{#1}
\csname url@rmstyle\endcsname
\providecommand{\newblock}{\relax}
\providecommand{\bibinfo}[2]{#2}
\providecommand\BIBentrySTDinterwordspacing{\spaceskip=0pt\relax}
\providecommand\BIBentryALTinterwordstretchfactor{4}
\providecommand\BIBentryALTinterwordspacing{\spaceskip=\fontdimen2\font plus
\BIBentryALTinterwordstretchfactor\fontdimen3\font minus \fontdimen4\font\relax}
\providecommand\BIBforeignlanguage[2]{{%
\expandafter\ifx\csname l@#1\endcsname\relax
\typeout{** WARNING: IEEEtran.bst: No hyphenation pattern has been}%
\typeout{** loaded for the language `#1'. Using the pattern for}%
\typeout{** the default language instead.}%
\else
\language=\csname l@#1\endcsname
\fi
#2}}

\bibitem{anderssonCasADiSoftwareFramework2019}
J.~A.~E. Andersson, J.~Gillis, G.~Horn, J.~B. Rawlings, and M.~Diehl, ``{{CasADi}}: A software framework for nonlinear optimization and optimal control,'' \emph{Mathematical Programming Computation}, Mar. 2019.

\bibitem{britoModelPredictiveContouring2019}
B.~Brito, B.~Floor, L.~Ferranti, and J.~{Alonso-Mora}, ``Model {{Predictive Contouring Control}} for {{Collision Avoidance}} in {{Unstructured Dynamic Environments}},'' \emph{IEEE Robotics and Automation Letters}, vol.~4, no.~4, pp. 4459--4466, Oct. 2019.

\bibitem{dubinsCurvesMinimalLength1957}
L.~E. Dubins, ``On {{Curves}} of {{Minimal Length}} with a {{Constraint}} on {{Average Curvature}}, and with {{Prescribed Initial}} and {{Terminal Positions}} and {{Tangents}},'' \emph{American Journal of Mathematics}, vol.~79, no.~3, pp. 497--516, 1957.

\bibitem{faulwasserModelPredictivePathfollowing2009}
T.~Faulwasser, B.~Kern, and R.~Findeisen, ``Model predictive path-following for constrained nonlinear systems,'' in \emph{Proceedings of the 48h {{IEEE Conference}} on {{Decision}} and {{Control}} ({{CDC}}) Held Jointly with 2009 28th {{Chinese Control Conference}}}, Dec. 2009, pp. 8642--8647.

\bibitem{friezeWorstcasePerformanceAlgorithms1982}
A.~M. Frieze, G.~Galbiati, and F.~Maffioli, ``On the worst-case performance of some algorithms for the asymmetric traveling salesman problem,'' \emph{Networks}, vol.~12, no.~1, pp. 23--39, 1982.

\bibitem{gruneNonlinearModelPredictive2017}
L.~Gr{\"u}ne and J.~Pannek, \emph{Nonlinear {{Model Predictive Control}}}, ser. Communications and {{Control Engineering}}.\hskip 1em plus 0.5em minus 0.4em\relax Cham: Springer International Publishing, 2017.

\bibitem{helsgaunEffectiveImplementationLin2000}
K.~Helsgaun, ``An effective implementation of the {{Lin}}--{{Kernighan}} traveling salesman heuristic,'' \emph{European Journal of Operational Research}, vol. 126, no.~1, pp. 106--130, Oct. 2000.

\bibitem{kaplanApproximationAlgorithmsAsymmetric2003}
H.~Kaplan, M.~Lewenstein, N.~Shafrir, and M.~Sviridenko, ``Approximation algorithms for asymmetric {{TSP}} by decomposing directed regular multigraphs,'' in \emph{44th {{Annual IEEE Symposium}} on {{Foundations}} of {{Computer Science}}, 2003. {{Proceedings}}.}, Oct. 2003, pp. 56--65.

\bibitem{lenyPerformanceOptimizationUnmanned2008}
J.~Le~Ny, ``Performance optimization for unmanned vehicle systems,'' Thesis, Massachusetts Institute of Technology, 2008.

\bibitem{lenyApproximationAlgorithmCurvatureConstrained2010}
J.~Le~Ny and E.~Feron, ``An {{Approximation Algorithm}} for the {{Curvature-Constrained Traveling Salesman Problem}},'' Oct. 2010.

\bibitem{lorenzenMPCbasedMotionPlanning2025}
M.~Lorenzen, T.~Alamo, M.~Mammarella, and F.~Dabbene, ``{{MPC-based}} motion planning for non-holonomic systems in non-convex domains,'' in \emph{European {{Control Conferences}}}, Thessaloniki, Greece, June 2025.

\bibitem{maRecedingHorizonPlanning2006}
X.~Ma and D.~A. Castanon, ``Receding {{Horizon Planning}} for {{Dubins Traveling Salesman Problems}},'' in \emph{Proceedings of the 45th {{IEEE Conference}} on {{Decision}} and {{Control}}}, Dec. 2006, pp. 5453--5458.

\bibitem{macharetSurveyRoutingProblems2018}
D.~G. Macharet and M.~F.~M. Campos, ``A survey on routing problems and robotic systems,'' \emph{Robotica}, vol.~36, no.~12, pp. 1781--1803, Dec. 2018.

\bibitem{macharetDataGatheringTour2012}
D.~G. Macharet, A.~A. Neto, V.~F. {da Camara Neto}, and M.~F.~M. Campos, ``Data gathering tour optimization for {{Dubins}}' vehicles,'' in \emph{2012 {{IEEE Congress}} on {{Evolutionary Computation}}}, June 2012.

\bibitem{mammarellaCooperationUnmannedSystems2022a}
M.~Mammarella, L.~Comba, A.~Biglia, F.~Dabbene, and P.~Gay, ``Cooperation of unmanned systems for agricultural applications: {{A}} theoretical framework,'' \emph{Biosystems Engineering}, vol. 223, pp. 61--80, Nov. 2022.

\bibitem{mehrezPredictivePathFollowing2017}
M.~W. Mehrez, K.~Worthmann, G.~K. Mann, R.~G. Gosine, and T.~Faulwasser, ``Predictive {{Path Following}} of {{Mobile Robots}} without {{Terminal Stabilizing Constraints}},'' \emph{IFAC-PapersOnLine}, vol.~50, no.~1, pp. 9852--9857, July 2017.

\bibitem{nyDubinsTravelingSalesman2012}
J.~Ny, E.~Feron, and E.~Frazzoli, ``On the {{Dubins Traveling Salesman Problem}},'' \emph{IEEE Transactions on Automatic Control}, vol.~57, no.~1, pp. 265--270, Jan. 2012.

\bibitem{rathinamResourceAllocationAlgorithm2007}
S.~Rathinam, R.~Sengupta, and S.~Darbha, ``A {{Resource Allocation Algorithm}} for {{Multivehicle Systems With Nonholonomic Constraints}},'' \emph{IEEE Transactions on Automation Science and Engineering}, vol.~4, no.~1, pp. 98--104, Jan. 2007.

\bibitem{rawlingsModelPredictiveControl2017}
J.~B. Rawlings, D.~Q. Mayne, and M.~Diehl, \emph{Model Predictive Control: Theory, Computation, and Design}, 2nd~ed.\hskip 1em plus 0.5em minus 0.4em\relax Madison, Wisconsin: Nob Hill Publishing, 2017.

\bibitem{savlaPointtopointTravelingSalesperson2005}
K.~Savla, E.~Frazzoli, and F.~Bullo, ``On the point-to-point and traveling salesperson problems for {{Dubins}}' vehicle,'' in \emph{Proceedings of the 2005, {{American Control Conference}}, 2005.}\hskip 1em plus 0.5em minus 0.4em\relax Portland, OR, USA: IEEE, 2005, pp. 786--791.

\bibitem{siegwartIntroductionAutonomousMobile2004}
R.~Siegwart, \emph{Introduction to Autonomous Mobile Robots}, ser. Intelligent Robots and Autonomous Agents.\hskip 1em plus 0.5em minus 0.4em\relax Cambridge, MA: MIT Press, 2004.

\bibitem{solopertoNonlinearMPCScheme2023}
R.~Soloperto, J.~K{\"o}hler, and F.~Allg{\"o}wer, ``A {{Nonlinear MPC Scheme}} for {{Output Tracking Without Terminal Ingredients}},'' \emph{IEEE Transactions on Automatic Control}, vol.~68, no.~4, pp. 2368--2375, Apr. 2023.

\bibitem{vanaOptimalSolutionGeneralized2020}
P.~V{\'a}{\v n}a and J.~Faigl, ``Optimal solution of the {{Generalized Dubins Interval Problem}}: Finding the shortest curvature-constrained path through a set of regions,'' \emph{Autonomous Robots}, vol.~44, no.~7, pp. 1359--1376, Sept. 2020.

\bibitem{wachterImplementationInteriorpointFilter2006}
A.~W{\"a}chter and L.~T. Biegler, ``On the implementation of an interior-point filter line-search algorithm for large-scale nonlinear programming,'' \emph{Mathematical Programming}, vol. 106, no.~1, pp. 25--57, Mar. 2006.

\bibitem{yuNonlinearModelPredictive2015}
S.~Yu, X.~Li, H.~Chen, and F.~Allg{\"o}wer, ``Nonlinear model predictive control for path following problems,'' \emph{International Journal of Robust and Nonlinear Control}, vol.~25, no.~8, pp. 1168--1182, 2015.

\end{thebibliography}
}

\end{document}